\documentclass[runningheads,a4paper]{llncs}
\usepackage[pass]{geometry}

\usepackage{graphicx}
\usepackage{epstopdf}
\usepackage[latin1]{inputenc}
\usepackage{amsfonts}
\usepackage{amssymb}
\usepackage{listings}
\usepackage{url}
\usepackage{algorithm2e}
\usepackage{gensymb}
\usepackage{textcomp}
\usepackage{caption}
\usepackage{subcaption}
\usepackage{wrapfig}
\usepackage[export]{adjustbox}

\usepackage[overlay,absolute]{textpos}

\begin{document}

\begin{textblock*}{10in}(38mm, 10mm)
{\textbf{Ref:} \emph{International Conference on Artificial Neural Networks (ICANN)}, Springer LNCS,}
\end{textblock*}
\begin{textblock*}{10in}(38mm, 15mm)
{Vol.~10614, pp.~287--296, Alghero, Italy, September, 2017.}
\end{textblock*}

\pagestyle{empty}

\author{Ido Cohen\inst{1} 
\and Eli (Omid) David\inst{1} 
\and Nathan S. Netanyahu\inst{1,2} 
\and \\Noa Liscovitch\inst{3}
\and Gal Chechik\inst{3}
}

\authorrunning{I. Cohen, O.E. David, N.S. Netanyahu, N. Liscovitch, G. Chechik}

\institute{
Department of Computer Science, Bar-Ilan University, Ramat-Gan, Israel \\
\email{cido15@gmail.com, mail@elidavid.com, nathan@cs.biu.ac.il}
\and
Center for Automation Research, University of Maryland, College Park, MD, USA\\
\email{nathan@cfar.umd.edu}
\and
Gonda Multidisiplinary Brain Research Center, Bar-Ilan University, Ramat-Gan, Israel\\
\email{noalis@gmail.com, gal.chechik@mail.biu.ac.il}
}

\title{\smaller DeepBrain: Functional Representation
of Neural In-Situ Hybridization Images
for Gene Ontology Classification\\ 
Using Deep Convolutional Autoencoders}

\titlerunning{DeepBrain: Functional Representation
of Neural In-Situ Hybridization Images
for Gene Ontology Classification 
Using Deep Convolutional Autoencoders}

\maketitle
\vspace{-10pt}

\begin{abstract}
This paper presents a novel deep learning-based method for learning a functional representation of mammalian neural images. The method uses a \textit{deep convolutional denoising autoencoder} (CDAE) for generating an invariant, compact representation of \textit{in situ hybridization} (ISH) images. While most existing methods for bio-imaging analysis were not developed to handle images with highly complex anatomical structures, the results presented in this paper show that functional representation extracted by CDAE can help learn features of functional \textit{gene ontology categories} for their classification in a highly accurate manner. Using this CDAE representation, our method outperforms the previous state-of-the-art classification rate, by improving the average AUC from 0.92 to 0.98, i.e., achieving 75\% reduction in error. The method operates on input images that were downsampled significantly with respect to the original ones to make it computationally feasible.

\end{abstract}

\section{Introduction}
\vspace{-6pt}

A very large volume of high-spatial resolution imaging datasets is available these days in various domains, calling for a wide range of exploration methods based on image processing. One such dataset has become recently available in the field of Neuroscience, thanks to the Allen Institute for Brain Science. This dataset contains \textit{in situ hybridization} (ISH) images of mammalian brains, in unprecedented amounts, which has motivated new research efforts \cite{Henry2012atlases}, \cite{Lein2007Genome}, \cite{Ng2009expression}.
ISH is a powerful technique for localizing specific nucleic acid targets within fixed tissues and cells; it provides an effective approach for obtaining temporal and spatial information about gene expression \cite{Puniyani2013GINI}.  Images now reveal highly complex patterns of gene expression varying on multiple scales.

However, analytical tools for discovering gene interactions from such data remain an open challenge due to various reasons, including difficulties in extracting canonical representations of gene activities from images, and inferring statistically meaningful networks from such representations. The challenge in analyzing these images is both in extracting the patterns that are most relevant functionally, and in providing a meaningful representation that allows neuroscientists to interpret the extracted patterns.

One of the aims at finding a meaningful representation for such images, is to carry out classification to \textit{gene ontology} (GO) categories. GO is a major Bioinformatics initiative to unify the representation of gene and gene product attributes across all species \cite{GO2008project}. More specifically, it aims at maintaining and developing a controlled vocabulary of gene and gene product attributes and at annotating them. This task is far from done; in fact, several gene and gene product functions of many organisms have yet to be discovered and annotated \cite{Ashburner2000GO}. Gene function annotations, which are associations between a gene and a term of controlled vocabulary describing gene functional features, are of paramount importance in modern biology. They are used to design novel biological experiments and interpret their results. Since gene validation through in vitro biomolecular experiments is costly and lengthy, deriving new computational methods and software for predicting and prioritizing new biomolecular annotations, would make an important contribution to the field \cite{Perez2004annotation}. In other words, deriving an effective computational procedure that predicts reliably likely annotations, and thus speed up the discovery of new gene annotations, would be very useful \cite{Plessis2011why}.

Past methods for analyzing brain images had to reference a brain atlas, and based on smooth non-linear transformations \cite{Davis2009Allen}, \cite{Hawrylycz2011Multi}. These types of analyses may be insensitive to fine local patterns, like those found in the layered structure of the cerebellum\footnote{The cerebellum is a region of the brain. It plays an important role in motor control, and has some effect on cognitive functions \cite{Wolf2009cerebellar}.}, or to spatial distribution. In addition, most machine vision approaches address the challenge of providing human interpretable analysis. Conversely, in bioimaging usually the goal is to reveal features and structures that are hardly seen even by human experts. For example, one of the new functions that follow this approach is presented in \cite{Liscovitch2013FuncISH}, using a histogram of local \textit{scale-invariant feature transform} (SIFT) \cite{Lowe2004SIFT} descriptors on several scales.

Recently, many machine learning algorithms have been designed and implemented to predict GO annotations \cite{Pinoli2015Computational}, \cite{Zitnik2014mold}, \cite{Vembu2014prediction}, \cite{Kordmahalleh2013Hierarchical}, \cite{King2013patterns}. In our research, we examine an \textit{artificial neural network} (ANN) with many layers (also known as \textit{deep learning}) in order to achieve functional representations of neural ISH images.

In order to find a compact representation of these ISH images, we explored \textit{autoencoders} (AE) and \textit{convolution neural networks} (CNN), and found the \textit{convolutional autoencoder} (CAE) to be the most appropriate technique. Subsequently, we use this representation to learn features of functional GO categories for every image, using a simple \textit{support vector machine} (SVM) classifier \cite{Vapnik1995svm}, as in \cite{Liscovitch2013FuncISH}. As a result, each image is represented as a point in a lower-dimensional space whose axes correspond to meaningful functional annotations. A similar example to ours is the work of Krizhevsky and Hinton \cite{Krizhevsky2011DeepAE}, who used deep autoencoders to create short binary codes for content-based images.
The resulting representations define similarities between ISH images which can be easily explained, hopefully, by such functional categories.

Our experimental results demonstrate that a so-called \textit{convolutional denoising autoencoder} (CDAE) representation (see Subsection~\ref{subsec:CAE}) outperforms the previous state-of-the-art classification rate, by improving the average AUC from 0.92 to 0.98, i.e., achieving 75\% reduction in error. The method operates on input images that were downsampled significantly with respect to the original ones to make it computationally feasible.

\section{Background}
\vspace{-6pt}

\subsection{FuncISH - Learning Functional Representations}

ISH images of mammalian brains reveal highly complex patterns of gene expression varying on multiple scales. Our study follows \cite{Liscovitch2013FuncISH}, which we pursue using deep learning. In \cite{Liscovitch2013FuncISH} the authors present \textit{FuncISH}, a learning method of functional representations of ISH images, using a histogram of local descriptors on several scales.

They first represent each image as a collection of local descriptors using SIFT features. Next, they construct a standard bag-of-words description of each image, giving a 2004-dimension representation vector for each gene. Finally, given a set of predefined GO annotations of each gene, they train a separate classifier for each known biological category, using the SIFT bag-of-words representation as an input vector. Specifically, they used a set of 2081 $L_2$-regularized logistic regression classifiers for this training. A scheme representing the work flow is presented in Figure~\ref{fig:flow} (see Section~\ref{sec:classification}).

Applying their method to the genomic set of mouse neural ISH images available from the Allen Brain Atlas, they found that most neural biological processes could be inferred from spatial expression patterns with high accuracy. Despite ignoring important global location information, they successfully inferred $\sim$700 functional annotations, and used them to detect gene-gene similarities which were not captured by previous, global correlation-based methods. According to \cite{Liscovitch2013FuncISH}, combining local and global patterns of expression is an important topic for further research, e.g., the use of more sophisticated non-linear classifiers.

\subsection{Deep Learning Techniques}
\vspace{-3pt}

Pursuing further the above classification problem poses a number of challenges. First, we cannot define a certain set of rules that an ISH image has to conform to in order to classify it to the correct GO category. Therefore, conventional computer vision techniques, capable of identifying shapes and objects in an image, are not likely to provide effective solutions to the problem. Thus, we use deep learning to achieve better results, as far as functional representations of the ISH images. This yields an interpretable measure of similarity between complex images that are difficult to analyze and interpret.

Deep learning techniques that support this kind of problems use AE and CNN, as well as CAE, which are successful in preforming feature extraction and finding compact representations for the kind of large ISH images we have been dealing with. While traditional machine learning is useful for algorithms that learn iteratively from the data, our second issue concerns the type of data we possess. Our data consist of 16K images, representing about 15K different genes, i.e., an average of one image per gene. This prevents us from extracting features from each gene independently, but rather consider the data in their entirety. Moreover, not only is there only one image per gene, there are merely a few genes in every examined GO category, and the genes are not unique to one category, i.e., each gene may belong to more than one category. Despite these difficulties, machine learning is capable of capturing underlying ``insights'' without resorting to manual feature selection. This makes it possible to automatically produce models that can analyze larger and more complex data, achieving thereby more accurate results. 

In the next section we present our convolutional autoencoder approach, which operates solely on raw pixel data. This supports our main goal, i.e., learning representations of given brain images to extract useful information, more easily, when building classifiers or other predictors. The representations obtained are vectors which can be used to solve a variety of problems, e.g., the problem of GO classification. 
For this reason, a good representation is also one that is useful as input to a supervised predictor, as it allows us to build classifiers for the biological categories known.

\section{Feature Extraction Using Convolutional Autoencoders}
\vspace{-6pt}

\subsection{Auto-Encoders (AE)}
\vspace{-3pt}
While convolutional neural networks (CNN) are effective in a supervised framework, provided a large training set is available, this is incompatible to our case.
If only a small number of training samples is available, unsupervised pre-training methods, such as \textit{restricted Boltzmann machines} (RBM) \cite{Hinton2006dbn} or autoencoders \cite{Vincent2008dae}, have proven highly effective.

An AE is a neural network which sets the target values (of the output layer) to be equal to those of the input, using hidden layers of smaller and smaller size, which comprise a bottleneck. Thus, an AE can be trained in an unsupervised manner, forcing the network to learn a higher-level representation of the input. 
An improved approach, which outperforms basic autoencoders in many tasks is due to \textit{denoising autoencoders} (DAEs) \cite{Vincent2008dae}, \cite{Vincent2010SDAE}. These are built as regular AEs, where each input is corrupted by added noise, or by setting to zero some portion of the values. Although the input sample is corrupted, the network's objective is to produce the original (uncorrupted) values in the output layer. Forcing the network to recreate the uncorrupted values results in reduced network overfitting (also due to the fact that the network rarely receives the same input twice), and in extraction of more high-level features. 
For any autoencoder-based approach, once training is complete, the decoder layer(s) are removed, such that a given input passes through the network and yields a high-level representation of the data. In most implementations (such as ours), these representations can then be used for supervised classification.

\subsection{Convolutional Autoencoders (CAE)}\label{subsec:CAE}
\vspace{-3pt}

\begin{figure}[ht]
	\centering
	\includegraphics[scale=0.5]{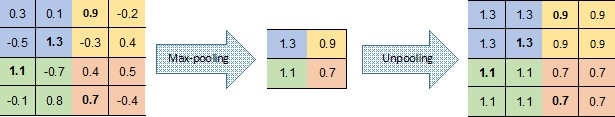}
	\caption{Pooling and unpooling layers; for each pooling layer, the max value is kept, and then duplicated in the unpooling layer.}
	\label{fig:pooling}
    \vspace{-10pt}
\end{figure}

CNNs and AEs can be combined to produce CAEs. 
As with CNNs, the CAE weights are shared among all locations in the input, preserving spatial locality and reducing the number of parameters.
In practice, to combine CNNs with AEs (or DAEs), it is necessary for each encoder layer to have a corresponding decoder layer. 
\textit{Deconvolution} layers are essentially the same as convolutional layers, and similarly to standard autoencoders, they can either be learned or set equal to (the transpose of) the original convolution layers, as with tied weights in autoencoders (both work well).
For the \textit{unpooling} operation, more than one method exists \cite{zeiler2014Visualizing}, \cite{masci2011scae}. In the CAE we use, during unpooling all locations are set to the maximum value which is stored in that layer (Figure~\ref{fig:pooling}).

Similarly to an AE, after training a CAE, the unpooling and deconvolution layers are removed. At this point, a neural net, composed from convolution and pooling layers, can be used to find a functional representation, as in our case, or initialize a supervised CNN. Similarly to a DAE, a CAE with input corrupted by added noise is called a \textit{convolutional denoising autoencoders} (CDAE).

\section{CDAE for GO Classification} \label{sec:classification}
\vspace{-6pt}

\begin{figure}[ht]
	\centering
	\includegraphics[width=1.02\textwidth,center]{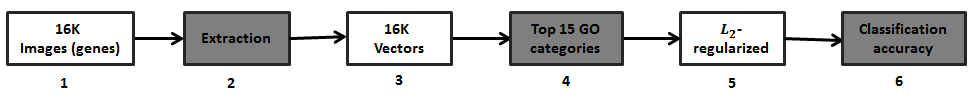}
	\caption{(1) 16K grayscale images indicating level of gene expression, (2) SIFT- or CDAE-based feature extraction for a compact vector representation of each gene, (3) vector representation due to feature extraction, (4) 16K vectors trained with respect to each of 15 GO categories with best AUC classification in \cite{Liscovitch2013FuncISH}, (5) $L_2$-regularized logistic regression classifiers for top 15 GO categories, and (6) classification accuracy measured.}
	\label{fig:flow}
\end{figure}

Figure~\ref{fig:flow} depicts a framework for capturing the representation of FuncISH. A SIFT-based module was used in \cite{Liscovitch2013FuncISH} for feature extraction. Alternatively, our scheme learns a CDAE-based representation, before applying a similar classification method as in \cite{Liscovitch2013FuncISH}, where two layers of 5-fold cross-validation were used, one for training the classifier and the other for tuning the logistic regression regularization hyperparameter.

For unsupervised training of our CDAE we use the genomic set of mouse neural ISH images available from the Allen Brain Atlas, which includes 16,351 images representing 15,612 genes. These JPEG images have an average resolution of $15,000 \times 7,500$ pixels. To get a representation vector of size $\sim$2,000, the images were downsampled to $300 \times 140$ pixels.

The CDAE architecture for finding a compact representation for these downsampled images is as follow: 
\big(1\big) \textbf{Input layer}: Consists of the raw image, resampled to $300 \times 140$  pixels, and corrupted by setting to zero 20\% of the values,
\big(2\big) three sequential convolutional layers with 32 $9 \times 9$ filters each,
\big(3\big) max-pooling layer of size $2 \times 2$,
\big(4\big) three sequential convolutional layers with 32 $7 \times 7$ filters each,
\big(5\big) max-pooling layer of size $2 \times 2$,
\big(6\big) two sequential convolutional layers with 64 $5 \times 5$ filters each,
\big(7\big) convolutional layer with a single $5 \times 5$ filter,
\big(8\big) unpooling layer of size $2 \times 2$,
\big(9\big) three sequential deconvolution layers with 32 $7 \times 7$ filters each,
\big(10\big) unpooling layer of size $2 \times 2$,
\big(11\big) three sequential deconvolution layers with 32 $9 \times 9$ filters each,
\big(12\big) deconvolution layer with a single $5 \times 5$ filter, and
\big(13\big) \textbf{output layer} with the uncorrupted resampled image.

After training the CDAE, all layers past item 8 are removed, so that item 7 (the convolutional layer of size $1 \times 2,625$) becomes the output layer. Therefore, each image is mapped to a vector of 2,625 functional features.
Given a set of predefined GO annotations for each gene (where each GO category consists of 15--500 genes), we trained a separate classifier for each biological category. Training requires careful consideration, in this case, due to the vastly imbalanced nature of the training sets. Similarly to  \cite{Liscovitch2013FuncISH}, we performed a weighted SVM classification using 5-fold cross-validation.

\medskip
This network yields remarkable AUC results for every category of the top 15 GO categories reported in \cite{Liscovitch2013FuncISH}.
Figure~\ref{fig:AUCplots} illustrates the AUC scores achieved for various representation vectors. While the average AUC score (of the top 15 categories) reported in \cite{Liscovitch2013FuncISH} was \textbf{0.92}, the average AUC using our CDAE scheme was \textbf{0.98}, i.e., a \textbf{75\%} reduction in error.

\subsection{Reducing Vector Dimensionality}
\begin{figure}[ht] 
\vspace{-15pt}
\begin{subfigure}{0.5\textwidth}
\includegraphics[scale=0.65]{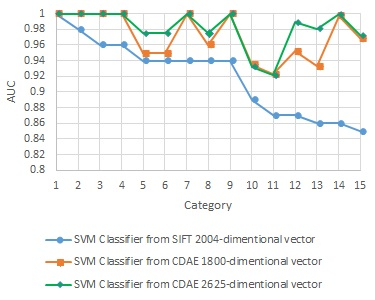}
\caption{}
\vspace{5pt}
\label{fig:sift_2625}
\end{subfigure}
{\hspace{1pt}}
\begin{subfigure}{0.5\textwidth}
\vspace{-5pt}
\includegraphics[scale=0.5]{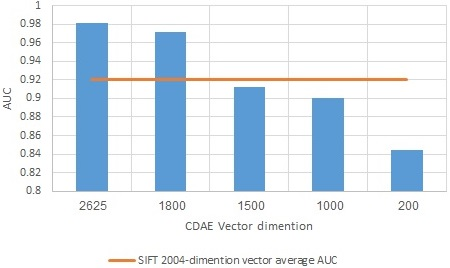}
\vspace{40pt}
\caption{}
\vspace{-6pt}
\label{fig:sizes}
\end{subfigure}
\caption{AUC results using convolutional denoising autoencoder for feature extraction: (a) AUC obtained from training a SVM classifier for each GO category, using a compact representation vector for every gene; representation vector dimensionality depends on method used and image resampling rate; (b) average AUC for top 15 classifiers, trained on different representation vectors due to CDAE (as in Figure~\ref{fig:cae}), for different resampling of brain images.}
\label{fig:AUCplots}
\vspace{-10pt}
\end{figure}

\vspace{-3pt}
The above improvement was achieved with a vector size of 2,625, which is larger than the 2004-dimensional vector obtained by SIFT.
In an attempt to maintain, as much as possible, the scheme's performance for a comparable vector size, we explored the use of smaller vectors, by resampling the images to different scales, and constructing CDAEs with various numbers of convolution and pooling layers. Figure~\ref{fig:AUCplots}(b) shows the average AUC for the top 15 categories mentioned earlier, with the same CDAE structure and the images resampled to smaller scales, thus obtaining lower-dimensionality representation vectors.

\medskip
Downsampling to $240 \times 120$ images, we obtained a 1800-dimensional representation vector, for which the AUC scores are still superior (relatively to \cite{Liscovitch2013FuncISH}) for each of the top 15 GO categories (as shown in Figure~\ref{fig:AUCplots}(a)). The \textbf{10\%}-dimensionality reduction results only in a slightly lower AUC average of \textbf{0.97} (see Figure~\ref{fig:AUCplots}(b)). 

\begin{figure}[ht]
  \vspace{15pt}	\includegraphics[width=1.02\textwidth,center]{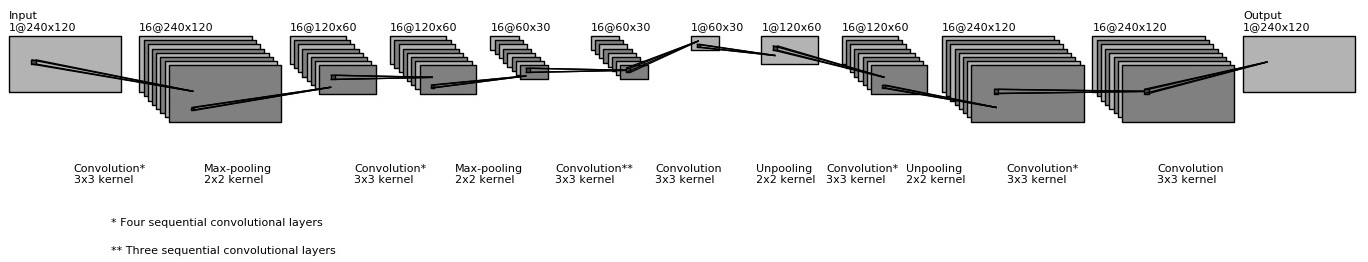}
      \vspace{-15pt}
	\caption{Illustration of our convolutional denoising autoencoder, achieving a compact representation for each gene.}
	\label{fig:cae}
    \vspace{-15pt}
\end{figure}

\medskip
The CDAE network for the more compact representation is shown in Figure~\ref{fig:cae}. The architecture consists of the following layers:
\big(1\big) \textbf{Input layer}: consists of the raw image, resampled to $240 \times 120$ pixels, and corrupted by setting to zero 20\% of the values,
\big(2\big) four sequential convolutional layers with 16 $3 \times 3$ filters each,
\big(3\big) max-pooling layer of size $2 \times 2$,
\big(4\big) four sequential convolutional layers with 16 $3 \times 3$ filters each,
\big(5\big) max-pooling layer of size $2 \times 2$,
\big(6\big) three sequential convolutional layers with 16 $3 \times 3$ filters each,
\big(7\big) convolutional layer with a single $3 \times 3$ filter,
\big(8\big) unpooling layer of size $2 \times 2$,
\big(9\big) four sequential deconvolution layers with 16 $3 \times 3$ filters each,
\big(10\big) unpooling layer of size $2 \times 2$,
\big(11\big) four sequential deconvolution layers with 16 $3 \times 3$ filters each,
\big(12\big) deconvolution layer with a single $3 \times 3$ filter, and
\big(13\big) \textbf{output layer} with the uncorrupted resampled image.

We used the ReLU activation function for all convolution and deconvolution layers, except for the last deconvolution layer, which uses tanh.

The learning rate starts from 0.05 and is  multiplied by 0.9 after each epoch, and the denoising effect is obtained by randomly removing 20\% of the pixels every image in the input layer.
We used the AUC as a measure of classification accuracy. 

\section{Conclusion}
\vspace*{-6pt}

Many machine learning algorithms have been designed lately to predict GO annotations. For the task of learning functional representations of mammalian neural images, we used deep learning techniques, and found convolutional denoising autoencoder to be very effective. Specifically, using the presented scheme for feature learning of functional GO categories improved the previous state-of-the-art classification accuracy from an average AUC of 0.92 to 0.98, i.e., a 75\% reduction in error. We demonstrated how to reduce the vector dimensionality by $10\%$ compared to the SIFT vectors, with very little degradation of this accuracy. Our results further attest to the advantages of deep convolutional autoencoders, as were applied here to extracting meaningful information from very high resolution images and highly complex anatomical structures. Until gene product functions of all species are discovered, the use of CDAEs may well continue to serve the field of Bioinformatics in designing novel biological experiments.

\section*{Appendix: Network Architecture Description}
\vspace{-6pt}

We provide a brief explanation as to the choice of the main parameters of the CDAE architecture. Our objective was to obtain a more compact feature representation than the 2,004-dimensional vector used in FuncISH.

Since a CNN is used, the representation along the grid should capture the two-dimensional structure of the input, i.e., the image dimensions should be determined according to the intended representation vector, while maintaining the aspect ratio of the original input image. Thus, we picked an 1,800-dimensional feature vector, corresponding to an (output) image of size $60 \times 30$. Taking into account the characteristic of max-pooling (i.e., that at each stage the dimension is reduced by 2), the desire to keep the number of layers as small as possible, and the fact that the encoding and decoding phases each contains the same number of layers (resulting in twice the number of layers in the network), we settled for two max-pooling layers, namely an input image of size $240 \times 120$.
Between each two max-pooling layers, which eliminate feature redundancy, there is an ``array'' of 16 convolution layers, each with the purpose of detecting locally connected features from its previous layer. The number of convolution layers (i.e., different filters used) was determined after experimenting with several different layers, all of which gave similar results. Choosing 16 layers (as shown in Figure~\ref{fig:cae}) provided the best result.

We experimented also with various filter sizes for each layer, ranging from
$3 \times 3$ to $11 \times 11$; while increasing the filter size significantly increased the amount of network parameters learned, it did not contribute much to the feature extraction or the improvement of the results.
Using a learning rate decay in the training of large networks (where there is a large number of randomly generated parameters) has proven helpful in the network's convergence. Specifically, the combination of a 0.05 learning rate parameter with a 0.9 learning rate decay resulted in an optimal change of the parameter value. In this case, too, small changes in the parameters did not result in significant changes in the results.

\bibliographystyle{plain}
\bibliography{DeepBrainBib}

\end{document}